\documentclass[letterpaper, 10 pt, conference]{ieeeconf}  

\IEEEoverridecommandlockouts                              

\usepackage{epsfig} 
\usepackage{mathptmx} 
\usepackage{times} 
\usepackage{amsmath} 
\usepackage{amssymb}
\usepackage{algorithm}
\usepackage{algpseudocode}
\usepackage{subfig}
\usepackage{booktabs}
\usepackage{array}
\usepackage{textcomp}
\usepackage{url}
\usepackage{verbatim}
\usepackage{hyperref}
\usepackage{graphicx}
\usepackage{cite}
\usepackage{makecell}
\usepackage{xcolor}
\newcommand{\Exp}{\mathbb{E}}

\DeclareMathOperator*{\argmin}{arg\,min} 
\DeclareMathOperator*{\rvec}{\mathrm{vec}}
\newtheorem{remark}{Remark}

\title{NANO-SLAM : \textbf{N}atural Gr\textbf{a}dient Gaussia\textbf{n} Appr\textbf{o}ximation for \\ Vehicle SLAM 
}

\author{Tianyi Zhang$^{1}$, Wenhan Cao$^{1}$, Chang Liu$^{2}$, Feihong Zhang$^{1}$, Wei Wu$^{1}$, Shengbo Eben Li$^{*}$
\thanks{$^{1}$Tianyi Zhang, Wenhan Cao, Feihong Zhang and Wei Wu are with the School of Vehicle and Mobility, Tsinghua University, Beijing, China (E-mail: \{zhantia24, cwh19, zfh24, w-wu18\}@mails.tsinghua.edu.cn).}
\thanks{$^{2}$Chang Liu is with the Department of Advanced Manufacturing
and Robotics, College of Engineering, Peking University, Beijing, China (e-mail: changliucoe@pku.edu.cn).
}
\thanks{$^{*}$Shengbo Eben Li is with the School of Vehicle and Mobility and College of Artificial Intelligence, Tsinghua University, Beijing, China (E-mail: lish04@gmail.com).}
\thanks{Corresponding Author: Shengbo Eben Li.}
}

\begin{document}

\maketitle
\thispagestyle{empty}
\pagestyle{empty}

\begin{abstract}
Accurate localization is a challenging task for autonomous vehicles, particularly in GPS-denied environments such as urban canyons and tunnels. In these scenarios, simultaneous localization and mapping (SLAM) offers a more robust alternative to GPS-based positioning, enabling vehicles to determine their position using onboard sensors and surrounding environment's landmarks. Among various vehicle SLAM approaches, Rao-Blackwellized particle filter (RBPF) stands out as one of the most widely adopted methods due to its efficient solution with logarithmic complexity relative to the map size. RBPF approximates the posterior distribution of the vehicle pose using a set of Monte Carlo particles through two main steps: sampling and importance weighting. The key to effective sampling lies in solving a distribution that closely approximates the posterior, known as the sampling distribution, to accelerate convergence. Existing methods typically derive this distribution via linearization, which introduces significant approximation errors due to the inherent nonlinearity of the system. To address this limitation, we propose a novel vehicle SLAM method called \textit{N}atural Gr\textit{a}dient Gaussia\textit{n} Appr\textit{o}ximation (NANO)-SLAM, which avoids linearization errors by modeling the sampling distribution as the solution to an optimization problem over Gaussian parameters and solving it using natural gradient descent. This approach improves the accuracy of the sampling distribution and consequently enhances localization performance. Experimental results on the long-distance Sydney Victoria Park vehicle SLAM dataset show that NANO-SLAM achieves over 50\% improvement in localization accuracy compared to the most widely used vehicle SLAM algorithms, with minimal additional computational cost.
\end{abstract}

\section{Introduction}

Autonomous vehicles represent a transformative technology with the potential to revolutionize transportation by improving efficiency, enhancing safety, and reducing accidents \cite{ShengboEbenLi}. For autonomous vehicles to navigate safely and make effective decisions, accurate localization is essential. In vehicle localization systems, GPS typically provides accurate positioning. However, in complex environments like urban canyons or underground areas, GPS signals may be weak, obstructed, or entirely unavailable \cite{levinson2010robust, liu2024particle, liu2021vehicle}.

To overcome this challenge, many systems leverage landmarks detected by onboard laser range finders as reference points, enabling vehicles to accurately compute their position relative to the environment \cite{engel2019deeplocalization, qu2017landmark}. This landmark-based approach is particularly useful in GPS-denied environments, making it a crucial component of advanced localization systems for autonomous vehicles. Nevertheless, in real-world scenarios, the landmark locations are typically unknown a priori, so both the landmark locations and the vehicle pose need to be estimated for localization \cite{thrun2002probabilistic}. From a probabilistic perspective, this problem involves solving the joint posterior distribution of the vehicle pose and all landmark locations, i.e., vehicle simultaneous
localization and mapping (SLAM) task. 

For this task, directly solving the joint posterior distribution is impractical for real-time operation in large-scale environments due to a computational cost that grows quadratically with the number of landmarks \cite{huang2007convergence}. Rao-Blackwellized particle filter (RBPF) \cite{murphy1999bayesian} is proposed to solve this issue by decomposing the joint posterior into the product of the vehicle pose posterior and the individual landmark location posterior, enabling separate estimation of the vehicle pose and landmark locations. When using a tree to store landmarks, the computational complexity of RBPF regarding the number of landmarks is logarithmic \cite{montemerlo2003fastslam}, which makes it more suitable for real-time applications. As a result, it is now a widely used landmark-based vehicle SLAM framework.
 
A representative RBPF-based algorithm is FastSLAM \cite{montemerlo2002fastslam}, which applies a standard particle filter (PF) to track the vehicle pose. In this approach, the vehicle pose in each particle is sampled based solely on the control input, without considering the current measurement, which often leads to suboptimal proposals, especially in scenarios with high measurement informativeness. FastSLAM 2.0 \cite{montemerlo2003fastslam} improves upon this by incorporating measurements into the vehicle pose sampling distribution, increasing the localization accuracy. However, it estimates the sampling distribution using first-order Taylor linearization for nonlinear systems, which introduces significant approximation errors. To overcome this limitation, Kim et al. proposed Unscented FastSLAM (UFastSLAM) \cite{kim2008unscented}, which replaces the first-order linearization in FastSLAM 2.0 with the unscented transformation (UT). The UT provides accuracy equivalent to a second-order Taylor expansion, thereby improving the approximation of nonlinear transformations.
To handle system nonlinearity more effectively, later works integrated methods such as spherical cubature integration  \cite{song2012square}  and transformed UT \cite{lin2018improved} into the RBPF framework to improve sampling quality and localization accuracy. Despite these advancements, all such methods fundamentally rely on linearization, which still introduce residual errors in highly nonlinear scenarios.
 
Inspired by a recent advancement in the Bayesian filtering field—the NANO filter \cite{cao2024nonlinear}, we propose \textit{N}atural Gr\textit{a}dient Gaussia\textit{n} Appr\textit{o}ximation (NANO)-SLAM, a novel RBPF-based vehicle SLAM algorithm that directly solves the vehicle pose sampling distribution using natural gradient descent, without relying on linear approximation. Specifically, our contributions are summarized as follows:

\begin{itemize}
    \item We formulate the solution to the vehicle pose sampling distribution as a variational optimization problem, avoiding the linearization process of the nonlinear system.
    \item To minimize the optimization objective, we cast the variational optimization problem as Gaussian parameter optimization and solve it via natural gradient iteration, which follows the steepest descent on the Gaussian manifold.

    \item We evaluate NANO-SLAM on the real-world Sydney Victoria Park dataset \cite{guivant2002simultaneous}. Experimental results show that our method achieves over a 50\% reduction in the root mean square error (RMSE) of vehicle localization compared to UFastSLAM, with minimal computational overhead.
\end{itemize}

\section{Preliminaries}
\label{sec:preliminaire}
In this section, we provide an introduction to the system model used for landmark-based vehicle localization, followed by a description of the RBPF framework commonly employed for this task, along with its associated challenges.
\subsection{System Modeling for Vehicle Localization}
\begin{figure}[t]
    \centering
\includegraphics[width=0.66\linewidth]{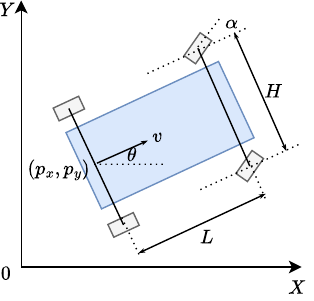}
    \caption{Vehicle in the 2D coordinate system.}
    \label{fig.vehicle}
\end{figure}

The autonomous vehicles
in the 2-dimensional (2D) coordinate system is shown in Fig. \ref{fig.vehicle}. In this figure, the coordinate $(p_x, p_y)$ represents the location of the vehicle, $\theta$ is the heading angle of the vechile; the vehicle velocity is denoted as $v$ and the steering angle is represented as $\alpha$. Besides, $L$ is the wheelbase and $H$ is the track width.

The discrete-time motion model of the vehicle follows the canonical Ackerman model as
\begin{equation}
\label{eq.motion_model}
\begin{aligned}
x_{t+1} &= f(x_t, u_t) + \xi_t
\\ 
&= x_t + \begin{bmatrix}
\Delta T \left( v_t \cos\theta_t - \frac{v_t}{L} \tan\theta_t (\alpha_t \sin\theta_t + b \cos\theta_t) \right) \\
\Delta T \left( v_t \sin\theta_t + \frac{v_t}{L} \tan\theta_t (\alpha_t \cos\theta_t + b \sin\theta_t) \right) \\
 \Delta T \frac{v_t}{L} \tan(\alpha_t) 
\end{bmatrix}
 + \xi_t.
\end{aligned}
\end{equation}
Here, the vehicle's pose consists of its location and heading angle, i.e., $x = [p_x, p_y, \theta]^{\top} \in \mathbb{R}^3$. Additionally, the control input consists of the vehicle's velocity and steering angle, i.e. $u = [v, \alpha]^{\top}\in \mathbb{R}^2$. The transition noise is assumed to be a zero mean Gaussian noise with covariance matrix $Q_t \in \mathbb{R}^{3\times3}$, i.e., $\xi_t \sim \mathcal{N}(0, Q_t)$. The time interval from consecutive time steps is denoted as $\Delta T$. Note that the velocity $v$ refers to the velocity of the center of the axle, which cannot be measured directly. Instead, it is transferred by the wheel velocity $v_e$ measured by the encoder located in the wheel. Typically, we have the following transformation rule
\begin{equation}
v = \frac{v_e}{ 1 -  H \cdot \tan{\alpha}/ (2L)}.
\end{equation}

Range-bearing measurements between landmarks and the autonomous vehicle are taken into account in this paper and can be obtained by using an onboard laser range finder. At each time step, the laser range finder provides $K$ measurements, \(z_t = \{z_{k,t}\}_{k=1}^K\) with \(z_{k,t}\in \mathbb{R}^2\), and the map contains $M$ landmarks, \(m_{1:M} =\{m_{j}\}_{j=1}^M \), where $m_j = \left[m_{x,j}, \, m_{y,j} \right]^\top \in \mathbb{R}^2$. 
Assuming that the $j$-th landmark corresponds to the $k$-th measurement at time $t$, then the measurement model can be represented as 
\begin{equation}
\label{eq.meas_model}
z_{k,t} = g(x_t, m_{j}) + \zeta_{k,t},     
\end{equation}
with
\begin{equation}
\begin{aligned}
\label{eq.meas_func}
g(x_t, m_{j}) =   \begin{bmatrix}
\sqrt{(m_{x,j} - p_{x,t})^2 + (m_{y,j} - p_{y,t})^2}\\
\arctan{\frac{m_{y,j} - p_{y,t}}{m_{x,j} - p_{x,t}}} - \theta_t
\end{bmatrix}
\end{aligned}.
\end{equation}
Here, $\zeta_{k,t} \in \mathbb{R}^{2}$ represents zero-mean Gaussian measurement noise with covariance matrix $R_t\in\mathbb{R}^{2\times2}$. 

\begin{remark}
The matching between $m_j$ and $z_k$ needs to be determined due to certain data associations. For example, the matching could be determined based on the Mahalanobis distance between the predicted measurement and the actual measurement.   
\end{remark}

\subsection{RBPF-based Vehicle Localization}
When using the measurement model \eqref{eq.meas_model} for vehicle localization, a key issue arises: the landmark locations $m_{1:M}$ are initially unknown. Therefore, to achieve vehicle localization, it is necessary to estimate the landmark locations as well. This is essentially a vehicle SLAM problem, where the objective is to estimate the joint posterior of the vehicle pose and all landmark locations as 
\begin{equation}
\label{eq.slam}
p\left(x_{1: t}, m_{1: M} \mid z_{1: t}, u_{0: t-1}, c_{1:t}\right).
\end{equation}
Here, $c_{1:t}$ represent the data association results, where each $c_t =\{(j,k)_i\}_{i=1}^C, C\leq M$, indicates $C$ landmarks have been matched to the measurements at time $t$, with the $j$-th landmark associated with the $k$-th measurement. The posterior \eqref{eq.slam} can be further factorized via Bayes' theorem \cite{montemerlo2002fastslam} into the product of the vehicle pose posterior and the posterior of each landmark's location, yielding
\begin{equation}
\label{eq.factorization}
\begin{aligned}
&p\left(x_{1: t}, m_{1: M} \mid z_{1: t}, u_{0: t-1}, c_{1:t}\right)\\
&= p\left(x_{1: t} \mid z_{1: t}, u_{0: t-1},c_{1:t}\right) p\left(m_{1: M} \mid x_{1: t}, z_{1: t},c_{1:t}\right) \\
&= p\left(x_{1: t} \mid z_{1: t}, u_{0: t-1},c_{1:t}\right) \prod_{i=1}^M p\left(m_i \mid x_{1: t}, z_{1: t},c_{1:t}\right).
\end{aligned}
\end{equation}

To solve the factorized posterior in \eqref{eq.factorization}, the RBPF framework is typically employed, which uses a particle set \(S_t = \{S_t^{(i)}\}_{i=1}^N\) to approximate this posterior distribution. Each particle is defined as
\begin{equation}
\label{eq.particles}
S_t^{(i)} = \left\{x_{t}^{(i)}, \mu_{1,t}^{(i)}, \Sigma_{1,t}^{(i)}, \dots, \mu_{M,t}^{(i)}, \Sigma_{M,t}^{(i)}\right\},
\end{equation}
where \(x_{t}^{(i)}\) represents the vehicle pose estimate, and \( \mu_{j,t}^{(i)} \) and \( \Sigma_{j,t}^{(i)} \) denote the mean and covariance of the Gaussian estimate for the \( j \)-th landmark location in the \( i \)-th particle. In each particle \(S_t^{(i)}\), the Gaussian estimate of each landmark's location can be independently solved using a 2D extended Kalman filter (EKF), and the vehicle pose estimate is obtained via sampling from the following distribution:
\begin{equation}
\label{eq.sample}
x_t^{(i)}\sim p(x_t|x_{1:t-1}^{(i)}, u_{0:t-1}, z_{1:t},c_{1:t}).
\end{equation}
For vehicle localization, the vehicle pose estimate is far more critical, thus accurately computing the sampling distribution is crucial.
 
Prior RBPF-based vehicle SLAM algorithms typically employed linearization \cite{montemerlo2003fastslam, kim2008unscented, lin2018improved} to solve the sampling distribution. However, these methods suffer from approximation errors due to the inherent nonlinearity of the system model \eqref{eq.motion_model}, \eqref{eq.meas_model}. Such inaccuracies in the sampling distribution adversely affect overall localization performance \cite{kim2008unscented}.To address this limitation, we propose NANO-SLAM in the following sections.

\section{NANO-SLAM Algorithm}
In this section, we will detail the proposed NANO-SLAM algorithm. Unlike previous methods that rely on linearization \cite{montemerlo2003fastslam, kim2008unscented, song2012square}, we formulate the vehicle pose sampling distribution as an optimization problem over the parameters of a Gaussian distribution, and solve it via natural gradient iteration, effectively avoiding linearization-induced errors. For landmark location estimation, we retain the use of EKF to strike a balance between accuracy and computational efficiency. Finally, importance sampling is applied to obtain the final vehicle localization result.

\subsection{NANO-Enhanced Vehicle Pose Estimation}
To solve the sampling distribution in \eqref{eq.sample}, we aim to find an approximating distribution \(q^{*}(x_t^{(i)})\) that is closest to this distribution. And the measure of this distance is usually chosen as the Kullback-Leibler (KL) divergence, which leads to the following variational problem:
\begin{equation}
\label{eq.variational}
\begin{aligned}
q^{*}(x_t) = \argmin_{q\in \mathcal{Q}}& \underbrace{D_{\mathrm{KL}}(q(x_t)\|p(x_t|x_{1:t-1}^{(i)}, u_{0:t-1}, z_{1:t},c_{1:t}))}_{J(q(x_t))},
\end{aligned}
\end{equation}
where $\mathcal{Q}$ denotes the admissible family of distributions, and
\begin{equation}
\label{eq.bayes}
\begin{aligned}
&p(x_t|x_{1:t-1}^{(i)}, u_{0:t-1}, z_{1:t},c_{1:t}) \\
&= \eta p(x_t|x_{1:t-1}^{(i)},u_{0:t-1}, z_{1:t-1},c_{1:t})p(z_t|z_{1:t-1}, x_t,x_{1:t-1}^{(i)},u_{0:t-1}, c_{1:t}) \\
&=\eta p(x_t|x_{t-1}^{(i)},u_{t-1})p(z_t| x_t, c_{t})
\end{aligned}
\end{equation}
Here, $\eta$ is a constant, \(p(x_t \mid x_{t-1}^{(i)}, u_{t-1})\) represents the prior sampling distribution and \(p(z_t|{x}_t, c_{t})\) denotes the measurement likelihood, which is also the probabilistic representation of the measurement model \eqref{eq.meas_model}. Substituting \eqref{eq.bayes} to \eqref{eq.variational} yields
\begin{equation}
\label{eq.loss}
\begin{aligned}
J(q(x_t)) =& D_{\mathrm{KL}}(q(x_t)\|p(x_t \mid x_{t-1}^{(i)}, u_{t-1})) \\
&+ \Exp_{q}(\log p(z_t\mid x_t, c_t)) + \mathrm{constant}    
\end{aligned}
\end{equation}

In general vehicle localization, the sampling distribution and its prior are typically assumed to be Gaussian \cite{chen2003bayesian, lin2018improved, zhang2024approximate}, as
\begin{equation}
\nonumber
\begin{aligned}
p(x_t \mid x_{t-1}^{(i)}, u_{t-1}) &= \mathcal{N}(x_t;\hat{x}_{t|t-1}^{(i)}, P^{(i)}_{t|t-1}), \\
p(x_t \mid x_{1:t-1}^{(i)}, u_{0:t-1}, z_{1:t},c_{1:t}) &= \mathcal{N}(x_t;\hat{x}_{t|t}^{(i)}, P^{(i)}_{t|t}),
\end{aligned}
\end{equation}
where $\hat{x}_{t|t-1}^{(i)}$ and $P_{t|t-1}^{(i)}$ denote the mean and covariance of the prior sampling distribution, which can be derived from \(q^{*}(x_{t-1})\) using the UT with the vehicle motion model in \eqref{eq.motion_model} \cite{kim2008unscented,zhang2024approximate}. Similarly, $\hat{x}_{t|t}^{(i)}$ and $P_{t|t}^{(i)}$ denote the mean and covariance of the sampling distribution. Thus, the approximating distribution can naturally be assumed to be Gaussian, as \(q(x_t^{(i)}) = \mathcal{N}(x_t;\hat{x}_t^{(i)}, P_t^{(i)})\). Substituting these Gaussian distributions into \eqref{eq.variational} and \eqref{eq.loss} reduces the variational problem to an optimization problem over Gaussian parameters, as
\begin{equation}
\begin{aligned}
\hat{x}_{t|t}^{(i)}, P_{t|t}^{(i)} =& \argmin_{\hat{x}_t^{(i)}, P_t^{(i)}} J(\hat{x}_t^{(i)}, P_t^{(i)}), \nonumber
\\
J(\hat{x}_t^{(i)}, P_t^{(i)}) =& D_{\mathrm{KL}}\left(\mathcal{N}(x_t; \hat{x}_t^{(i)}, P_t^{(i)}) || \mathcal{N}(x_{t}; \hat{x}_{t|t-1}^{(i)}, P_{t|t-1}^{(i)})\right)   \nonumber
\\
&-\Exp_{\mathcal{N}(x_t; \hat{x}_t^{(i)}, P_t^{(i)})} \big\{\log{p(z_t|{x}_t, c_t)}\big\} \\
=&  \frac{1}{2} \left(\hat{x}_{t|t-1}^{(i)}-\hat{x}_t^{(i)} \right)^{\top} (P_{t|t-1}^{(i)})^{-1}\left(\hat{x}_{t|t-1}^{(i)}-\hat{x}_t^{(i)} \right)
\\
&+ \frac{1}{2} \mathrm{Tr}\left((P_{t|t-1}^{(i)})^{-1} P_t^{(i)} \right)
-\frac{1}{2} \log \frac{\left|P_t^{(i)}\right|}{\left|P_{t|t-1}^{(i)}\right|} \\
&-\Exp_{\mathcal{N}(x_t; \hat{x}_t^{(i)}, P_t^{(i)})} \big\{\log{p(z_t|{x}_t, c_t)}\big\}.
\end{aligned}  
\end{equation}
To solve this parameter optimization problem, we derived a natural gradient iteration method that can find the steepest descent direction in the Gaussian parameter space \cite{amari1998natural,martens2020new}.

For simplicity, we stack the Gaussian parameters into a single column vector $v$ and calculate the derivative with respect to it:
\begin{equation}
\nonumber
\begin{aligned}
v &= \begin{bmatrix}
\hat{x}_t^{(i)} \\ \rvec((P_t^{(i)})^{-1})
\end{bmatrix}, \\ \nabla_{v} J &= \begin{bmatrix}
\frac{\partial}{\partial \hat{x}_t^{(i)}} J(\hat{x}_t^{(i)}, P_t^{(i)})  \\ \rvec\left(\frac{\partial}{\partial (P_t^{(i)})^{-1}} J(\hat{x}_t^{(i)}, P_t^{(i)})\right)
\end{bmatrix}.
\end{aligned}
\end{equation}
The update quantity for the natural gradient iteration is defined as
\begin{equation}
\Delta v =  -\mathcal{F}^{-1}_v\nabla_{v} J,
\end{equation}
where \(\mathcal{F}^{-1}_v\) is the inverse of the Fisher information matrix, specified as
\begin{equation}\label{eq.fisher matrix inverse}
\begin{aligned}
\mathcal{F}^{-1}_v = 
\begin{bmatrix}
P_t^{(i)} & 0 \\
0 & 2 ((P_t^{(i)})^{-1} \otimes (P_t^{(i)})^{-1})
\end{bmatrix}, 
\end{aligned}   
\end{equation}
and $\otimes$ represents the Kronecker product. Therefore, the $r$-th natural gradient iteration can be expressed as
\begin{equation}\label{eq.natural gradient update 1}
\begin{aligned}
 (P_t^{(i)})^{-1}_{(r+1)} =& (P_{t|t-1}^{(i)})^{-1}  + \Exp_{\mathcal{N}(x_t; \hat{x}_{t,(r)}^{(i)}, P_{t,(r)}^{(i)})} \left\{ \frac{\partial^2 \ell(x_t, z_t, c_t)}{\partial x_t^2} \right\},\\
\hat{x}^{(i)}_{t,(r+1)} =& \hat{x}^{(i)}_{t,(r)} - P_{t,(r+1)}^{(i)} 
\Exp_{\mathcal{N}(x_t; \hat{x}_{t,(r)}^{(i)}, P_{t,(r)}^{(i)})} \left\{ \frac{\partial \ell(x_t, z_t, c_t)}{\partial x_t} \right\} \\
&-
P_{t,(r+1)}^{(i)} (P_{t|t-1}^{(i)})^{-1}(\hat{x}_{t,(r)}^{(i)} - \hat{x}_{t|t-1}^{(i)}).
\end{aligned}  
\end{equation}
Here, $\ell(x_t, z_t, c_t)$ is defined as the negative of the log-likelihood:
\begin{equation}
\nonumber
\begin{aligned}
\ell(x_t, z_t, c_t) =& -\log p(z_t|x_t,c_t)\\
=& \sum_{(j,k)\in c_t}\frac{1}{2}(z_{k,t} - g(x_t, \mu_{j,t-1}))^\top R_t^{-1}(z_{k,t} - g(x_t, \mu_{j,t-1})) \\
&+ \mathrm{constant},
\end{aligned}
\end{equation}
and thus its first and second derivatives can be derived as 
\begin{equation}
\label{eq.derivate}
\begin{aligned}
\frac{\partial \ell(x_t, z_t, c_t)}{\partial x_t} &= -\sum_{(j,k)\in c_t}G_{x,t}^\top R^{-1}(z_{k,t} - g(x_t, \mu_{j,t-1})),\\
\frac{\partial^2 \ell(x_t, z_t, c_t)}{\partial x_t^2} &\approx  \sum_{(j,k)\in c_t}G_{x,t}^\top R^{-1}G_{x,t},
\end{aligned}
\end{equation}
where $G_{x,t} = \left.\frac{\partial g}{\partial x}\right|_{x=x_t}$ denotes the measurement Jacobian matrix with respect to the vehicle pose, evaluated
at \(x=x_t\). The approximation of the second derivative in \eqref{eq.derivate} is commonly used in various optimization methods \cite{foresee1997gauss}. 
Substituting \eqref{eq.derivate}  into \eqref{eq.natural gradient update 1} and perform the natural gradient iteration until
\begin{equation}
\nonumber
D_{\mathrm{KL}}(\mathcal{N}(x_t;\hat{x}^{(i)}_{t,(r)} ,P_{t,(r)}^{(i)}) \| \mathcal{N}(x_t;\hat{x}^{(i)}_{t,(r+1)} ,P_{t,(r+1)}^{(i)}) < \gamma,
\end{equation}
where $\gamma$ is a predefined threshold. Then, the mean and covariance of the Gaussian sampling distribution at time $t$ are obtained as
\begin{equation}
\nonumber
\hat{x}^{(i)}_{t|t}=\hat{x}^{(i)}_{t,(r+1)},\quad P_{t|t}^{(i)}=P_{t,(r+1)}^{(i)}.
\end{equation}
Finally, the vehicle pose estimate in the particle \eqref{eq.particles} is derived as
\begin{equation}
\nonumber
\hat{x}^{(i)} \sim \mathcal{N}(x_t; \hat{x}^{(i)}_{t|t}, P_{t|t}^{(i)}).
\end{equation}

\subsection{EKF for landmark locations estimation}
In landmark-based vehicle localization, the accuracy of landmark location estimation has a relatively minor effect on performance. Hence, to balance estimation accuracy with computational efficiency, the EKF is retained for landmark location estimation.

Specifically, in the data association results $c_t$, if a measurement \(z_{k,t}\)  is not associated with an existing landmark, a new landmark needs to be initialized as
\begin{equation}
\label{eq.init}
\mu_{\mathrm{new},t}^{(i)} = g^{-1}(\hat{x}_{t|t}^{(i)}, z_{k,t}), \quad \Sigma_{\mathrm{new},t}^{(i)} = G_{\mathrm{new},t}^{-1}RG_{\mathrm{new},t}^{-\top},
\end{equation}
where \(G_{\mathrm{new},t} \triangleq \left. \frac{\partial g(\hat{x}_{t|t}^{(i)}, m)}{\partial m} \right|_{m = \mu_{\mathrm{new},t}}\) is the measurement Jacobian matrix with respect to the landmark location, evaluated at \(m = \mu_{\mathrm{new},t}^{(i)}\). If the measurement is associated with an existing landmark whose location estimate has mean \(\mu_{j,t-1}^{(i)}\) and covariance \(\Sigma_{j,t-1}^{(i)}\), the EKF is used to update the landmark location as
\begin{equation}
\nonumber
\begin{aligned}
 K_{j,t}^{(i)} &= \Sigma_{j,t-1}^{(i)}{G_{j,t}}^{\top}(G_{j,t}\Sigma_{j,t-1}^{(i)}G_{j,t}^{\top}+R_t)^{-1}, \\    
 \mu_{j,t}^{(i)} &= \mu_{j,t-1}^{(i)} + K_{j,t}^{(i)}(z_{k,t} - g(\hat{x}_{t|t}^{(i)}, \mu_{j,t-1}^{(i)})),\\
\Sigma_{j,t}^{(i)} &= (I - K_{j,t}^{(i)}{G_{j,t}})\Sigma_{j,t-1}^{(i)}.
\end{aligned}
\end{equation}

\subsection{Importance Sampling}
In the RBPF framework, after obtaining each particle's vehicle pose and landmark location estimates, importance sampling is required to derive the final vehicle localization result. Importance sampling begins by computing the importance weight of each particle \cite{montemerlo2003fastslam, kim2008unscented}, defined as
\begin{equation}
\label{eq.weight}
\begin{aligned}
w_t^{(i)} &= \frac{\mathrm{target \ distribution}}{\mathrm{proposal \ distribution}}\\
&=\frac{p(x_t^{(i)}|z_{1:t},u_{1:t-1}, c_{1:t})}{p(x_{1:t-1}^{(i)}|z_{1:t-1},u_{1:t-2}, c_{1:t-1})p(x_t^{(i)}|x_{1:t-1}^{(i)},z_{1:t},u_{1:t-1}, c_{1:t})}\\
&=\sum_{(j,k)\in c_t}|2 \pi L_{j,t}^{(i)}|^{-\frac{1}{2}} \exp \left\{-\frac{1}{2}(z_{k,t}-\hat{z}_{k,t}^{(i)})^T {L_{j,t}^{(i)}}^{-1}(z_{k,t}-\hat{z}_{k,t}^{(i)})\right\},
\end{aligned}
\end{equation}
where 
\begin{equation}
\nonumber
\hat{z}_{k,t}^{(i)} = g(\hat{x}_{t|t}^{(i)}, \mu_{j,t}^{(i)}), \quad L_{j,t}^{(i)} = {G_{\hat{x}^{(i)},t}} P_{t|t}^{(i)}{G_{\hat{x}^{(i)},t}}^\top + G_{j,t} \Sigma_{j,t}^{(i)}{G_{j,t}}^\top + R_t.
\end{equation}
Note that the second term in the proposal distribution of~\eqref{eq.weight} corresponds to the vehicle pose sampling distribution. Thus, improving its accuracy enhances the overall proposal distribution, which in turn boosts localization performance.
Finally, based on the computed importance weights, the vehicle pose of the particle with the highest weight is selected as the final localization result.

\begin{figure}[t]
    \centering
\includegraphics[width=0.7\linewidth]{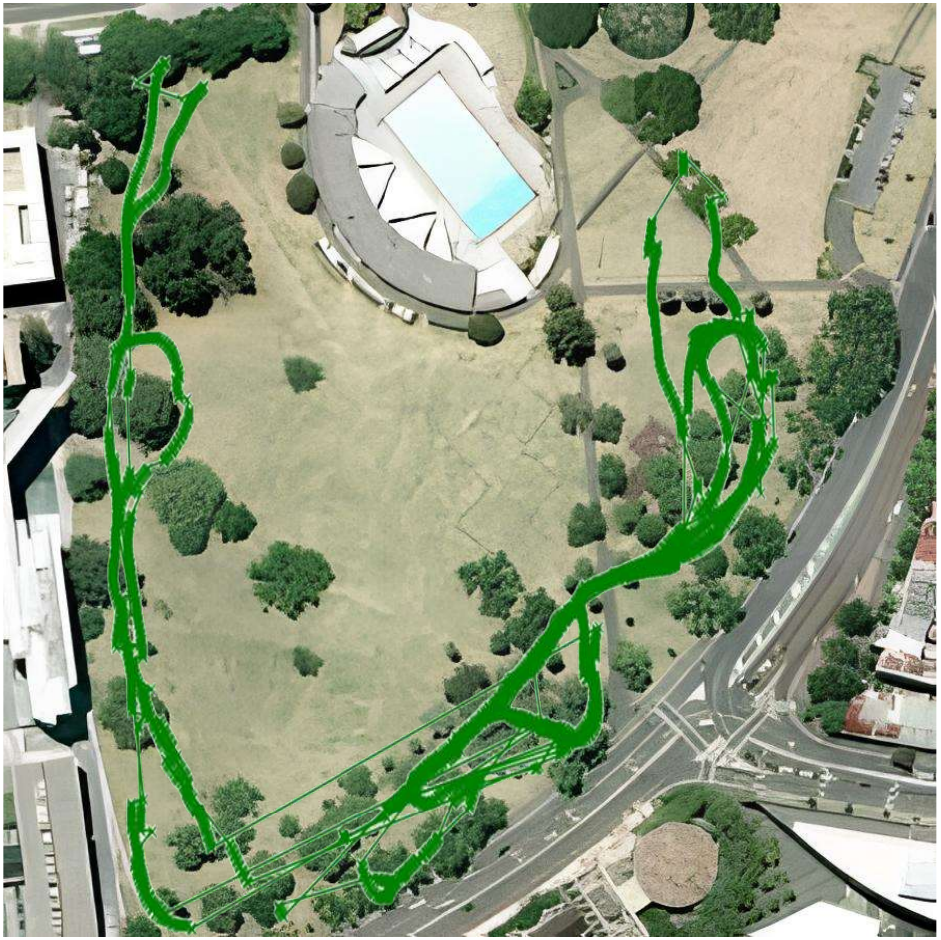}
    \caption{Victoria Park map \cite{guivant2002simultaneous} and vehicle trajectory obtained from GPS data.}
    \label{fig.map}
\end{figure}

\section{Experimental Results}
\begin{figure}[!t]
    \centering
    \subfloat[]{
        \includegraphics[width=0.45\textwidth]{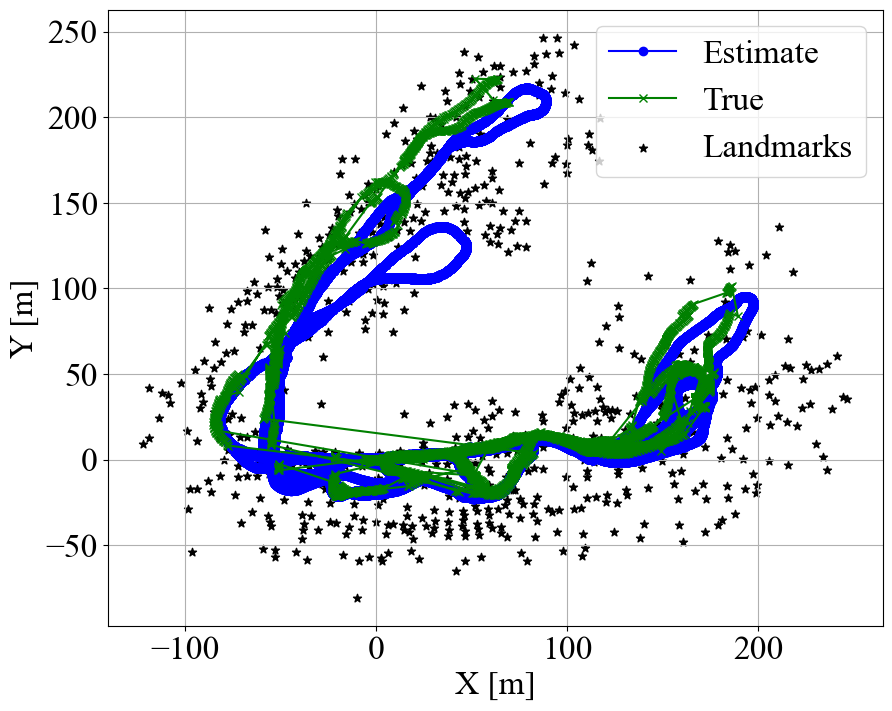}
        \label{est:1}
    }
    \\
    \subfloat[]{
        \includegraphics[width=0.45\textwidth]{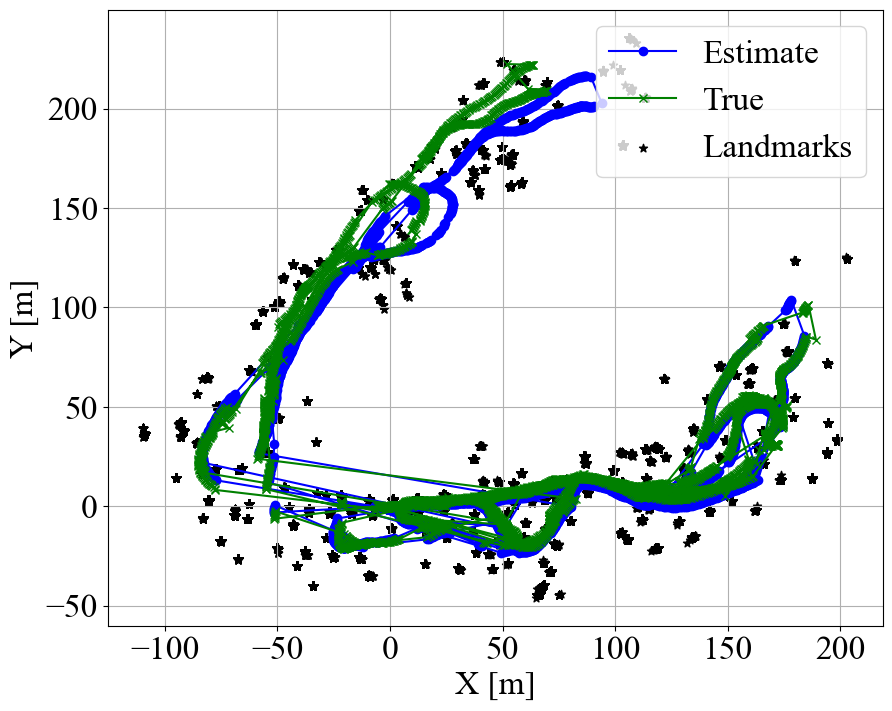}
        \label{est:2}
    }
    \\
    \subfloat[]{
        \includegraphics[width=0.45\textwidth]{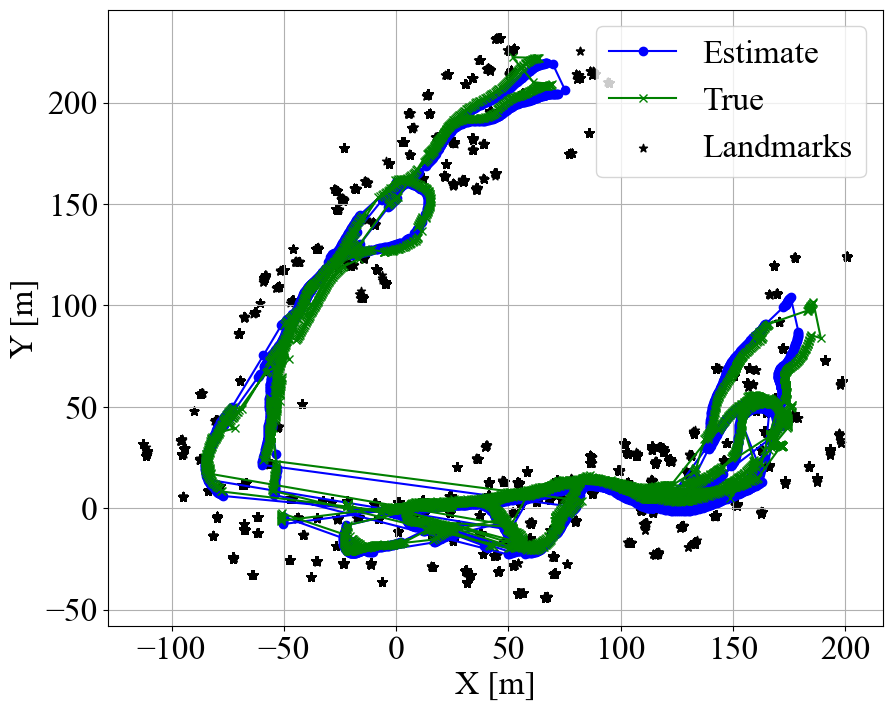}
        \label{est:3}
    }
    \caption{Experimental results of the vehicle localization and landmark location estimates. (a) EKF-SLAM. (b)UFastSLAM. (c)NANO-SLAM}
    \label{fig.estimate}
\end{figure}
To assess the effectiveness of NANO-SLAM in large-scale, real-world scenarios, we conduct experiments using the Sydney Victoria Park dataset. In this dataset, a vehicle equipped with wheel encoders, a laser range finder, and GPS traverses a park environment, as illustrated in Fig.~\ref{fig.map}, covering a trajectory of more than 3.5 kilometers. The wheel encoders provide measurements of the vehicle’s velocity and steering angle, which serve as the control inputs. The laser range finder covers a 180-degree frontal field of view and provides range-bearing measurements for landmark detection, with nearby trees serving as natural landmarks. The GPS is used to provide the ground-truth position of the vehicle, which is then used to compute localization errors. The vehicle has a wheelbase of 2.83 m and a track width of 0.76 m.

\begin{figure}[t]
    \centering
\includegraphics[width=0.76\linewidth]{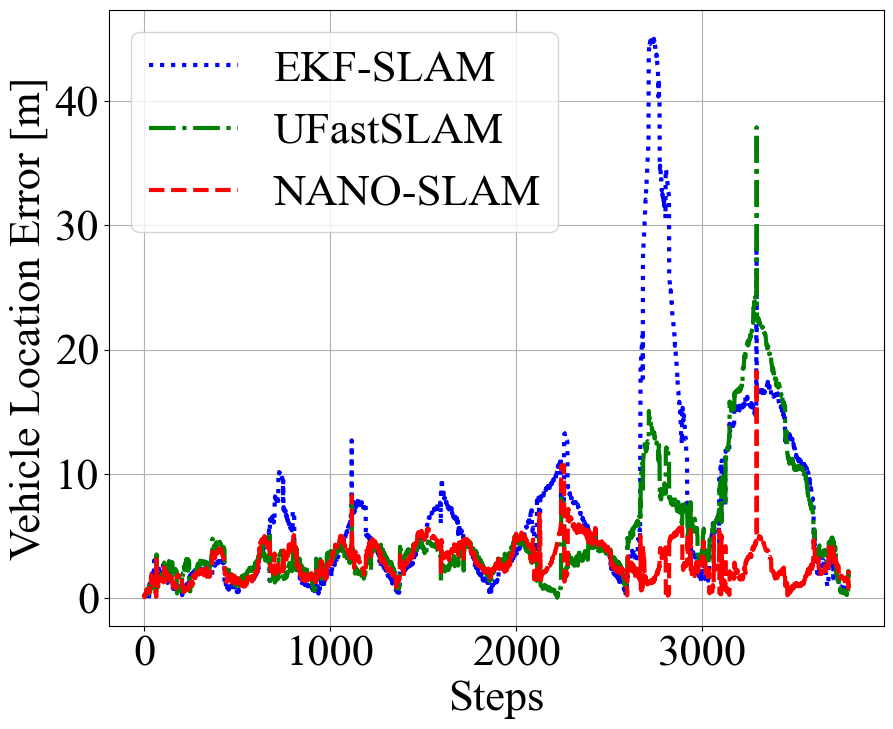}
    \caption{Vehicle localization errors over time.}
    \label{fig.error}
\end{figure}

\begin{figure}[t]
    \centering
\includegraphics[width=0.76\linewidth]{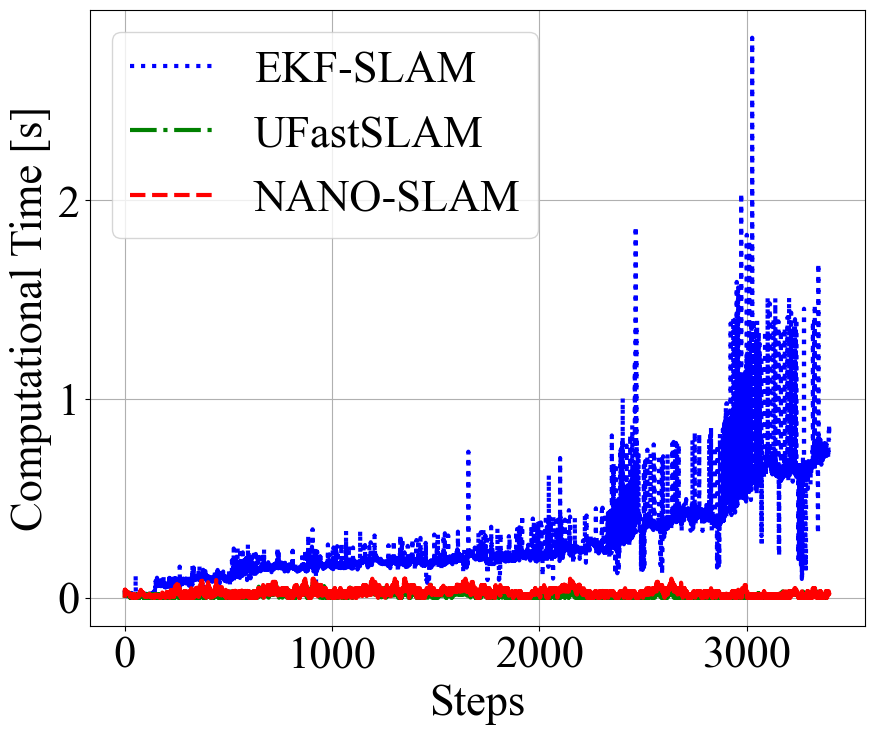}
    \caption{Computational Time of different methods.}
    \label{fig.time}
\end{figure}

In our experiments, we compare the proposed NANO-SLAM with EKF-SLAM \cite{huang2007convergence} and UFastSLAM \cite{kim2008unscented}, which are representative of direct solution-based and RBPF-based approaches to vehicle SLAM, respectively. To account for non-systematic errors such as irregular terrain and wheel slippage, the standard deviations of velocity and steering angle noises are set to $\sigma_v = 2$ m/s and $\sigma_g = 6$ deg, respectively, while the range and bearing measurement noise are modeled with $\sigma_r = 1$ m and $\sigma_b = 3$ deg to reflect sensor and environmental uncertainty. Ten particles are used in UFastSLAM and NANO-SLAM. Localization accuracy is evaluated by the root mean square error (RMSE), defined as
\begin{equation}
\label{eq.rmse}
\mathrm{RMSE} = \sqrt{\frac{1}{T}\sum_{t=1}^T\|p_t - \hat{p}_t\|_2^2},
\end{equation}
where $p_t$ and $\hat{p}_t$ denote the ground-truth and estimated vehicle positions at time step $t$, and $T$ is the total number of time steps.

\begin{table}[htbp]
    \centering
    \caption{Comparison of different SLAM Methods.}
    \label{tab:slam_comparison}
    \begin{tabular}{lcc}
        \toprule
        \textbf{Method} & \textbf{RMSE [m]} & \textbf{Time [ms]} \\
        \midrule
        EKF-SLAM    & 7.783 & 312.79 \\
        UFastSLAM   & 5.147 & \textbf{14.896} \\
        NANO-SLAM   & \textbf{2.538} & 17.692 \\
        \bottomrule
    \end{tabular}
\end{table}

As shown in Table.~\ref{tab:slam_comparison}, the proposed NANO-SLAM reduces the RMSE by over 50\% compared to UFastSLAM, the second-best method, by avoiding nonlinear approximation errors in the measurement model, with only a 18\% loss in computational efficiency. Fig.~\ref{fig.estimate} shows the estimated vehicle trajectory and landmark locations, where the proposed NANO-SLAM yields the trajectory that best matches the ground truth, demonstrating superior localization accuracy. In addition, Fig.~\ref{fig.error} presents the localization error over time. NANO-SLAM maintains the lowest and most stable error throughout the trajectory, whereas UFastSLAM and EKF-SLAM exhibit larger fluctuations in error as the vehicle progresses. Finally, Fig.~\ref{fig.time} illustrates the per-step computation time, which increases steadily for EKF-SLAM due to its quadratic complexity from full covariance updates as the state dimension grows, while remaining nearly constant for both UFastSLAM and NANO-SLAM thanks to their more efficient landmark management with only logarithmic complexity.

\section{Conclusion}
In this paper, we present NANO-SLAM, an efficient RBPF-based vehicle SLAM algorithm that models the vehicle pose sampling distribution as the solution to an optimization problem, and solves it using the natural gradient. This approach effectively eliminates the impact of linear approximation errors, improving sampling accuracy without sacrificing computational efficiency. Experimental validation on the Sydney Victoria Park dataset demonstrates that our method significantly outperforms existing vehicle SLAM algorithms, achieving over a 50\% improvement in localization accuracy with negligible additional computational cost. These results highlight the potential of NANO-SLAM to enable more precise and robust localization for real-world autonomous driving applications.

\bibliographystyle{IEEEtran}
\bibliography{ref}

\end{document}